\pdfoutput=1
\documentclass[10pt, a4paper]{article}
\usepackage{lrec2026}

\usepackage{silence}
\usepackage[T1]{fontenc}
\usepackage[utf8]{inputenc}
\usepackage{microtype}
\usepackage{amsmath}
\usepackage{booktabs}
\usepackage{graphicx}
\usepackage{multirow}
\usepackage{sansmath}
\sansmath
\usepackage{tabularx}
\title{%
    Predicting States of Understanding in Explanatory Interactions Using Cognitive Load-Related Linguistic Cues}

\name{%
    Yu Wang$^{\dagger\|}$ \: 
    Olcay Türk$^{\dagger\|}$ \:
    Angela Grimminger$^{\ddagger\|}$ \:
    Hendrik Buschmeier$^{\dagger\|}$ 
}

\address{%
  $^\dagger$Faculty of Linguistics and Literary Studies, Bielefeld University, Bielefeld, Germany\\
  $^\ddagger$Faculty of Arts and Humanities, Paderborn University, Paderborn, Germany\\
  $^\|$SFB/Transregio 318 ‘Constructing Explainability’, Bielefeld \& Paderborn, Germany
}

\abstract{%
    We investigate how verbal and nonverbal linguistic features, exhibited by speakers and listeners in dialogue, can contribute to predicting the listener's state of understanding in explanatory interactions on a moment-by-moment basis. Specifically, we examine three linguistic cues related to cognitive load and hypothesised to correlate with listener understanding: the information value (operationalised with surprisal) and syntactic complexity of the speaker's utterances, and the variation in the listener's interactive gaze behaviour. Based on statistical analyses of the MUNDEX corpus of face-to-face dialogic board game explanations, we find that individual cues vary with the listener's level of understanding. Listener states (`Understanding', `Partial Understanding', `Non-Understanding' and `Misunderstanding') were self-annotated by the listeners using a retrospective video-recall method. The results of a subsequent classification experiment, involving two off-the-shelf classifiers and a fine-tuned German BERT-based multimodal classifier, demonstrate that prediction of these four states of understanding is generally possible and improves when the three linguistic cues are considered alongside textual features.
    \\\vspace{-2mm}\\\Keywords{%
        dialogue, understanding, cognitive load, gaze, information value, syntactic complexity
    }
}

\begin{document}

\maketitleabstract

\section{Introduction}
\label{sec:introduction}

Explanatory interactions are a type of everyday communicative activity in which an `explainer' tries to explain something to an `explainee'. The explainee seeks to understand the explanation and common ground is continually built during the interaction \citep{ClarkBrennan1991}. In the grounding process, explainees frequently provide feedback, either verbally (e.g., in form of backchannels) or nonverbally (e.g., in form of head gestures), to explainers, and by that display or signal different states of understanding \citep[e.g.,][]{Allwoodetlal1992, AllwoodKopp2007}: understanding might be signalled through nodding, partial understanding by more hesitant nodding, and non-understanding by providing feedback with decreased volume of the voice. Feedback allows explainers to monitor an explainee's state of understanding and potentially adapt their explanation such that mutual understanding can be co-constructed \citep{ClarkKrych2004}. The social practice of explaining entails processes such as monitoring understanding of the explainee and adapting to the explainee. These processes have recently been proposed as important components in building social explainable AI systems \citep[‘Social XAI’;][]{RohlfingCimiano2021.TCDS, rohlfing2026sxai}. 

Monitoring understanding moment-by-moment requires identification of the (non-)verbal cues that reflect different states of understanding throughout the interaction. Thus, understanding -- and, more specifically, difficulties in understanding that might increase cognitive load -- may be signalled by certain verbal and non-verbal behaviour. Previous studies show that backchannels such as `mhm' or `ja' are efficient vocal signals that communicate addressees' understanding \citep{Allwoodetlal1992}. Non-verbal signal such as gaze, can indicate cognitive processing activity, as supported by empirical experiments in which participants solve challenging tasks \citep{Glenbergetal1998}. As noted by \citet{turk2024predictability}, variation in gaze direction demonstrates the extent of cognitive processing effort and can thus indicate an explainee's level of understanding -- to a certain degree. Furthermore, surprisal theory \citep{hale2001} posits that the cognitive effort (cognitive load) required to process language depends on its contextual predictability, quantified as ‘surprisal’. This was later substantiated numerically with empirical and theoretical evidence from reading studies \citep[e.g.,][]{LEVY20081126, smith2013effect, shain-2021-cdrnn}.

In this paper, we hypothesise that explainee's understanding is related to cognitive load \citep[e.g.,][]{languagesbetz}. Following this hypothesis, we review previous studies to identify potential verbal and nonverbal cues that can indicate cognitive load. From these, we select gaze variation, information value, and syntactic complexity as potential correlates. We use MUNDEX, a multimodal corpus of explanation dialogues \citeplanguageresource{Turk2023MUNDEX}, in which the above-mentioned cues are well-represented in annotated behaviour of explainees. We quantify these cues, first performing statistical analyses to verify their significance for explainees' understanding. Based on these analyses, we then use selected linguistic cues to perform a classification experiment on explainees' states of understanding.

Our main contributions in this paper are:
    (1) We investigate three linguistic (verbal and nonverbal) cues that previous linguistic studies considered as important indicators of cognitive load in language comprehension. Based on these cues that are also considered to be related to the explainees' understanding states, we obtain four different values to represent cognitive load, three of which we find to significantly vary with explainees' states of understanding.
    (2) We use two off-the-shelf classifiers and trained a BERT-based classifier to investigate the predictability of different states of understanding. Unlike previous work, which predicted understanding using a binary classification task (understanding vs. non-understanding) based on a series of multimodal signals \citep{KinoshitaOnishi2023, turk2024predictability}, we quantify linguistic cues potentially indicating cognitive load and perform multi-class classification of different understanding states. Results show that all of the classifiers perform better than chance, which indicates the feasibility of predicting different understandings states. Among the three classifiers, the BERT-based classifier is most robust. The variation in performance when predicting different understanding labels also suggests potential challenges for future work.

\section{Background}
\label{sec:background}

\subsection{Linguistic Cues for Cognitive Load}
\label{sec:bg:lingcue}

Cognitive load is considered as the amount of working memory dedicated to problem solving \citep{SWELLER1988257, Paas01012003}. Language comprehension, like many other daily tasks, constantly requires working memory capacity \citep{Just1992-JUSACT}. From a psycholinguistic perspective, cognitive load, sometimes conceptualised as ‘processing difficulty’ \citep{LEVY20081126, mitchell-etal-2010-syntactic, languagesbetz}, is relevant for various aspects of language processing. One example is syntactic complexity, which denotes the cognitive load to parse and process a sentence or an utterance \citep{szmrecsanyi2004operationalizing}.
Dependency locality theory \citet{GIBSON19981, gibson2000dependency} suggests that language comprehension involves continuous integration cost for processing sentences with varied syntactic structures. Accordingly, the theory treats syntactic complexity as a potential indicator of cognitive load. This is supported by empirical evidence and analyses showing that sentences that rank as more syntactically complex are considered more difficult for humans to process \citep{lin1996structural}, or that specific syntactic components, such as nouns or specific type of verbs, induce higher processing difficulty compared to other syntactic components \citep{DEMBERG2008193, demberg2009computational}.

In language comprehension, cognitive load can also be estimated by the predictability of a word in its context. This is called ‘surprisal’ \citep{hale2001, LEVY20081126}. Given a word $w$ and its context, e.g., a sentence that comprises of a sequence of smaller units: 
    $\langle w_1, \ldots, w_i \rangle$,
where $w_i \in \vartheta$, with $\vartheta$ being the vocabulary, surprisal of the word is modelled as the negative log probability: 
    $-\log P(w_i \mid w_1, \ldots, w_{i-1})$.
The higher the surprisal value of a word, the more unpredictable it is, and the greater the cognitive load required to process it. Surprisal, as well as its related concept such as contextual entropy, are widely accepted as models of the effort for language comprehension, given their psychometric predictive power on cognitive load (measured, e.g., through reading time; see \citealp[]{frank2013uncertainty,wilcox2023testing}). Focusing on dialogic interaction, previous studies show that surprisal based information values converge and diverge continuously while a dialogue unfolds \citep{xu2018information}. This indicates a potential correlation between information value and interlocutors' development of understanding \citep{maes2022shared}.

In addition to text based linguistic cues, nonverbal cues have also been discussed in terms of their relation to cognitive load. One example is gaze. In face to face interaction, the gaze behaviour of listeners serves as critical cues indicating their visual attention to the speaker's ongoing verbal content \citep{kendon1967some}. Moreover, gaze aversion is considered to be evidence of cognitive processing when answering questions \citep{Glenbergetal1998}, but has also been described as part of the so-called “thinking face” in interactional settings, displaying language processing \citep[e.g.,][]{Bavelas_Chovil_2018}. Empirical studies further show that gaze aversion is more likely to occur when there is an increase of cognitive load \citep{morency2006recognizing, Glenbergetal1998}. The variation of gaze during language comprehension in interaction can thus be considered a potential indicator of an interlocutors' cognitive load.

\subsection{Computational Modelling for Understanding Evaluation}
\label{sec:bg:modelling-understanding}

Evaluation of a human interlocutor's understanding is an emerging but important task in human interaction with adaptive conversational agents: conversational agents, such as embodied conversational agents, social robots, but also LLM-based chatbots, should be able to estimate the understanding of their interlocutor in order to adjust their utterances and by that, promote efficient communication \citep{reidsma2011continuous, BuschmeierKopp2018, axelsson2022multimodal, robrecht2023snape, mindlin2024measuring}. A competent conversation agent should also be able to correct any misunderstanding of the interlocutors when misunderstanding occurs. Regarding the evaluation and prediction of understanding in interaction, \citet{HowesEshghi2021} model the effect of interlocutors' backchannels by incrementally tracking them as evidence of understanding. Similarly, \citet{BuschmeierKopp2018} propose a probabilistic model in which an agent uses multimodal signals of the interlocutor to represent the current grounding state and update its belief about the listener's state of understanding during the interaction.

More recent work treats the evaluation of understanding as a machine learning task. \citet{KinoshitaOnishi2023}, for example, built a dialogue corpus consisting of the listener's comprehension levels (a digital value from $-2$ to $2$, annotated by third-party annotators) and the listener's multimodal information, and ran a regression model to predict the comprehension level. Similarly, \citet{turk2024predictability} investigate the predictability of understanding in the MUNDEX corpus \citeplanguageresource{Turk2023MUNDEX}, with understanding labels created based on participants' self-report of their understanding during the game explanation \citep{lazarov2025applications}.

\section{Dataset: MUNDEX Corpus}
\label{sec:dataset}

\begin{figure}
    \includegraphics[width=\linewidth]{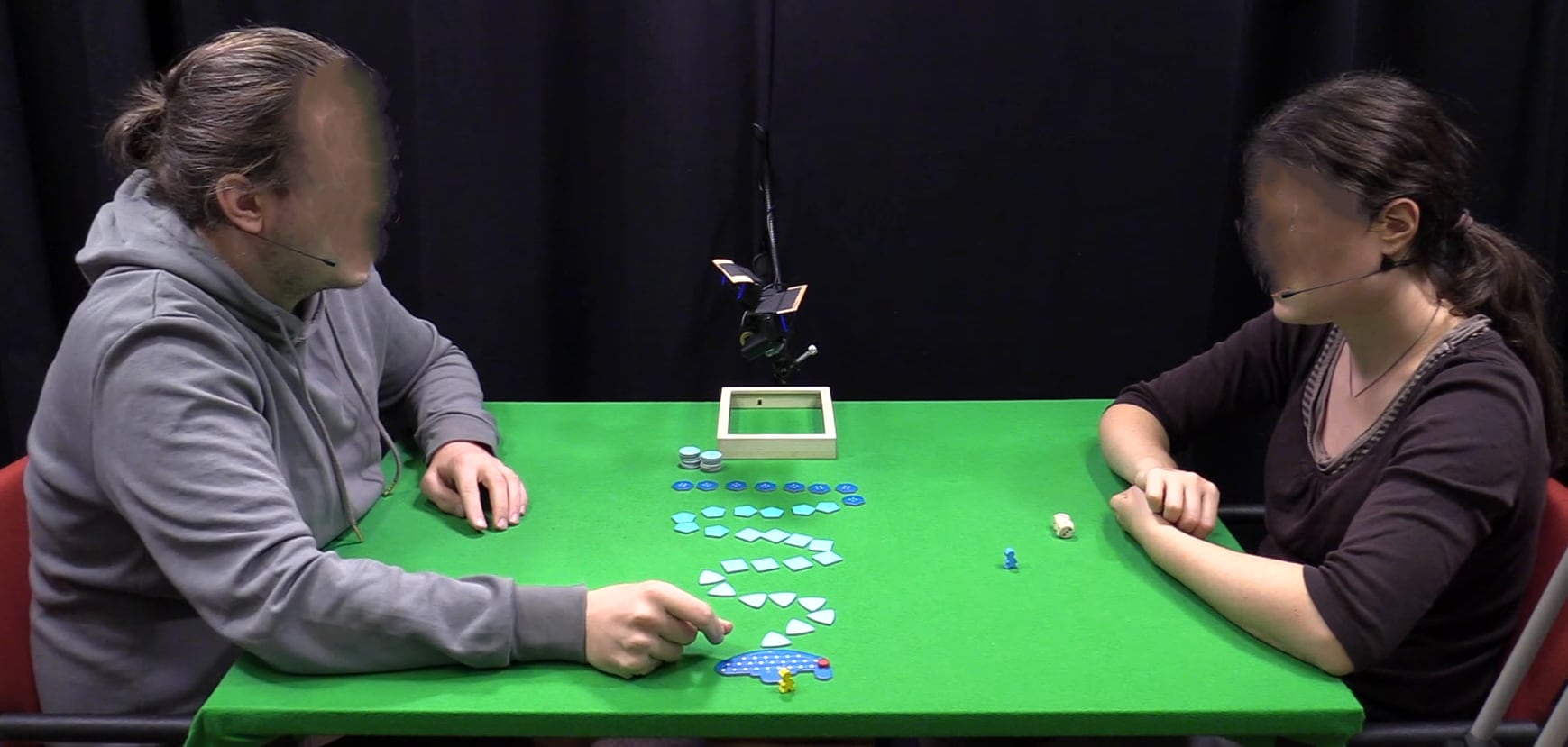}
    \caption{%
        Explanation set-up in the MUNDEX corpus (screenshot from one camera-perspective). The person on the left is the explainer who explains a board game; the person on the right is the explainee.}
    \label{fig:placeholder}
\end{figure}

For our research, we used a the MUNDEX corpus 
\citeplanguageresource{Turk2023MUNDEX} involving dyadic explanations of how to play a board game (Figure~\ref{fig:placeholder}). MUNDEX contains manual and automatic multimodal annotations \citeplanguageresource{buschmeier2025mundex} of the acoustic signal (e.g., voice quality), textual descriptors (e.g., discourse functions), and nonverbal behaviour (e.g., gaze, head gestures, and adaptors). The corpus was created to study how different states of understanding of explanations are multimodally signalled. These states were annotated using ‘retrospective video-recall’, a self-annotation method \citep{lazarov2025applications} in which
explainees watched a recording of their own interaction (immediately after the interaction was completed) and commented on their state of understanding (into four levels). These comments were grouped under two labels: understanding and non-understanding. Inter-annotator agreement regarding understanding annotations was high (Cohen's $\kappa = 0.90$; \citealp{turk2024predictability}).

In the current study, we expand on the analysis of \citet{turk2024predictability}, using these four levels of understanding. The additional understanding labels reflect the interaction dynamics more deeply, such as when the explainee only understood part of the content (i.e., Partial Understanding), or they understood what was explained in a different way (i.e., Misunderstanding). This expansion enables a more comprehensive investigation of understanding, acknowledging its gradual nature.
Here, we selected a total of 21 explanatory dialogues from MUNDEX and calculated the distribution of different understanding labels (Table~\ref{tab:understanding-labels}): Understanding (29.8\%) and Non-Understanding (27.4\%) have slightly higher proportions compared to Partial Understanding (25.2\%) and Misunderstanding (17.6\%).

\begin{table}
\small
\begin{tabularx}{\linewidth}{lXX}
\toprule
    \textbf{Label}        & \textbf{N} & \textbf{\%} \\
\midrule
    Understanding         & 176        & 27.4 \\ 
    Partial Understanding & 162        & 25.2 \\ 
    Non-Understanding     & 191        & 29.8 \\ 
    Misunderstanding      & 113        & 17.6 \\ 
\bottomrule
\end{tabularx}
\caption{%
    Distribution of labels for different states of understanding extracted from annotations during the game explanation phase.}
\label{tab:understanding-labels}
\end{table}

\begin{figure*}
    \centering
    \includegraphics[width=\linewidth]{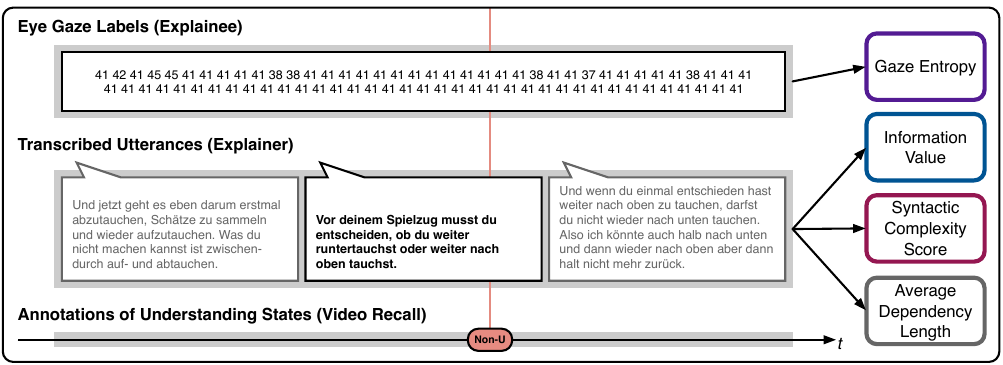}
    \caption{%
        The general quantification pipeline for getting average information value, average gaze entropy, average syntactic complexity score, and average dependency length.}
    \label{fig:pipe}
\end{figure*}

\section{Methods}
\label{sec:methods}

In Section~\ref{sec:bg:lingcue}, we discussed potential linguistic cues, indicating cognitive load during language comprehension, to analyse. Based on the survey, we choose the following three linguistic aspects in order to see how effective they are for predicting different understanding states:
    (i) semantic information conveyed in the utterances (measured using information value);
    (ii) structural complexity of the utterances (measured using syntactic complexity score and dependency length); and
    (iii) variation in the listener's gaze behaviour (measured using gaze entropy).
In the following, we introduce the methods for cognitive load quantification.

\paragraph{Information Value Quantification}

We quantified the information value of utterances based on the work of \citet{xu2018information} and \citet{giulianelli2021information}, where the information value of an utterance, $H(X)$, is defined as the average surprisal of each word in the utterance. Given an utterance of words  $\langle w_1, \dots, w_{N} \rangle$, average gaze entropy is defined as:
\begin{equation*}
    H(X) = -\frac{1}{N} \sum_{i = 1}^{N}\log P(w_i \mid w_{<i})
\end{equation*}
The greater the information value, the more unpredictable the content. In our study, being exposed to utterances with higher information value is expected to increase the explainee's cognitive load. We measure utterance surprisal by estimating the information content with GPT-2 \citep[specifically, dbmdz/german-gpt2;][]{german-gpt}.

\medskip
For our approach to syntactic complexity quantification of utterances, we now introduce Syntactic Complexity Score and Average Dependency Length as two alternative metrics.

\paragraph{Syntactic Complexity Score}

Given a speaker's utterance, we calculate its length $L$ and use dependency parsing \citep{qi2019universal} to obtain the number of heads $\alpha$ as well as the maximum tree depth $\beta$ \citep[see][]{wang-buschmeier-2024-revisiting}.
The syntactic complexity $SC$ of the utterance is then computed as follows:
\begin{equation*}
    SC = 
    \begin{cases}
        \lambda \cdot \frac{L}{\alpha} + (1 - \lambda) \cdot \beta  
            & \text{if } \alpha > 0\\
        (1 - \lambda) \cdot \beta  
            & \text{otherwise,}
    \end{cases}
\end{equation*}
where $\lambda$ is a tuning factor ($0.5$ by default). The formula above takes the following aspects into consideration: Generally
    (i) an utterance with longer length ($L$) requires more working memory, and thus a higher cognitive load; 
    (ii) an utterance with more heads ($\alpha$) in its syntactic structure is harder to process; 
    (iii) an utterance with larger tree depth ($\beta$) tends to have more embedded clauses, which is also a sign of syntactic complexity.

\paragraph{Average Dependency Length}

An alternative approach for quantifying syntactic complexity relies on the average dependency length. Dependency length minimization theory states that speakers will minimise dependency length in order to decrease processing difficulty \citep{futrell2015dependency}, which indicates that the dependency length is a decent measurement for syntactic complexity.
We follow \citet{liu2008dependency} and \citet{futrell2015dependency} to formally define dependency length as follows: We regard an utterance as a large syntactic structure. Let an utterance of length $n$ be a syntactic structure where each node will have one arc. Its representation can then have a set of dependencies 
    $D = \{(h_i, d_i)\}_{i=1}^n$,
where each $(h_i, d_i)$ denotes a syntactic dependency between a head word at position $h_i$ and its dependent at position $d_i$. The dependency length for a word and its dependent is the number of words in between. The average dependency length of an utterance then represents its syntactic complexity:
\begin{equation*}
    \mathrm{ADL} = \frac{1}{|D|} \sum_{(h, d) \in D} |h - d|
\end{equation*}
In general, a larger average dependency length value can indicate that the utterance is more lengthy and has more embedded syntactic structure, and thus, requires more working memory to process.

\paragraph{Average Gaze Entropy}

Gaze variation during interaction is considered an important nonverbal cue to estimate a listener's understanding. To quantify gaze variation, we use the method proposed in \citet{wang-buschmeier-2023-listener} and \citet{wang-etal-2024-much}. It quantizes the speech aligned gaze vector, generated from explainee videos using OpenFace \citep{baltrusaitis2018openface}, into a discrete number ranging from $1$ to $81$, where $41$ is the most frequent label indicating explainee's gaze direction is on the explainer (attention on the explainer). The other values suggest gaze aversion to varying degrees.  Figure~\ref{fig:pipe} illustrates how the explainee's gaze is aligned with the words of the explainer and quantized into a gaze labels (the numbers). Next, we apply a transformer model that has been trained using gaze label data in order to obtain gaze entropy values. Given a sequence of gaze labels  $\langle e_1, \dots, e_{T} \rangle$, the average gaze entropy is defined as:
\begin{equation*}
    \mbox{NLL}(e_1, e_2, \ldots, e_T) 
        = - \frac{1}{T} \sum_{i=1}^{T}\log P(e_i \mid e_{<i})
\end{equation*} 
The higher the average gaze entropy value, the more unpredictable the gaze label becomes. This can indicate different degrees of gaze variation. Variation in gaze, such as for example gaze aversion -- considered a cue for memory search \citep{Glenbergetal1998} -- is expected when potentially high cognitive load is present on the side of the explainee.

\paragraph{Quantification Pipeline}

The schema in Figure~\ref{fig:pipe} illustrates the pipeline we use to obtain the quantifications for our statistical analysis and understanding state classification. For each annotated understanding state in our corpus (`Non-Understanding' in the example shown in Figure~\ref{fig:pipe}), we first identify its corresponding utterance. This utterance is then extended with its immediate context in form of its preceding and its succeeding utterance. In Figure~\ref{fig:pipe}, the utterance corresponding with the annotated understanding state is shown in the second box (in bold), while the preceding and succeeding utterances (in the first and third boxes, in grey) are prepended and appended to form its context. This processing is based on our observation that understanding states can be confirmed with greater confidence when contextual factors are taken into consideration. The combined utterances are then used to compute the average information value, syntactic complexity score, and average dependency length value. In addition, they are aligned with the corresponding gaze labels of the explainee (gaze behaviour while these utterances were spoken) to compute the average gaze entropy value.

\section{Results and Discussion}
\label{sec:results}

In our survey of previous studies, we discussed linguistic cues which may be related to the development of understanding during interaction. Here, we examine selected linguistic cues from the MUNDEX corpus and quantify the linguistic information based on the approach proposed in Section~\ref{sec:methods}. We first present a statistical analysis of the individual cues, which is followed by classification experiments involving the cues.\footnote{%
    All values used in the statistical analyses and classification tasks have been normalised using the function \emph{MinMaxScaler} from scikit-learn \citep{pedregosa2011scikit}.
}

\subsection{Statistical Analysis}
\label{sec:stat}

\begin{figure*}
    \includegraphics[width=1\linewidth]{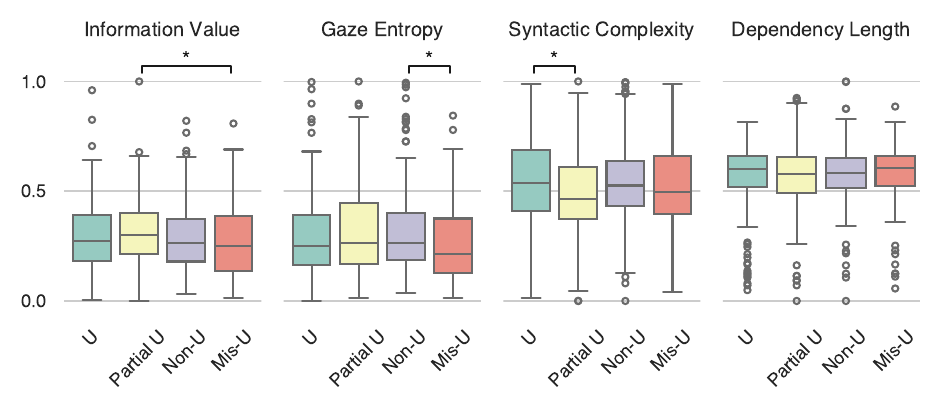}
    \vspace{-9mm}
    \caption{%
        Variation of the quantified linguistic cues under different states of understanding (‘U’). Horizontal bars show statistically significant Dunn's post-hoc tests (Bonferroni-corrected, $\alpha = 0.05$).}
    \label{fig:boxplots}
\end{figure*}

With our statistical analysis we aim to establish whether the selected linguistic cues individually vary among the four different states of understanding. As can be seen by comparing the boxplot diagrams in Figure~\ref{fig:boxplots}, the median \textit{information value} is lower for misunderstanding and non-understanding, the median \textit{gaze entropy} is lower for misunderstanding, and the median \textit{syntactic complexity} score is higher in understanding and non understanding. For dependency length, the medians are almost identical across the four different understanding states.

For our statistical analysis, we first conduct Kruskal-Wallis tests for each linguistic cue, assuming no difference in median value among the understanding states as the null hypothesis. The results, as shown in Table~\ref{tab:kruskal-wallis-stats}, indicate statistically significantly different medians between understanding states for information value, gaze entropy, and syntactic complexity scores, but not for average dependency length ($\alpha=0.05$). It should be noted though that effect sizes ($\eta^2$) are very small.

\begin{table}
\small
\begin{tabularx}{\linewidth}{lXXl}
    \toprule
    \textbf{Predictor} & \textbf{H} & \textbf{\textit{p}} & \textbf{$\eta^2$} \\
    \midrule
        Information Value    & $9.029$ & $0.0289$ & $ 0.0094$ \\
        Gaze Entropy         & $9.035$ & $0.0288$ & $ 0.0095$ \\
        Syntactic Complexity & $8.853$ & $0.0313$ & $ 0.0092$ \\
        Dependency Length    & $1.344$ & $0.7187$ & $-0.0025$ \\
    \bottomrule
\end{tabularx}
\caption{%
    Kruskal-Wallis test results showing associations between different understanding states and each predictor variable.}
    \label{tab:kruskal-wallis-stats}
\end{table}

These analyses are then followed up by pairwise post-hoc tests between understanding states (Dunn's test with Bonferroni correction for multiple comparison). Results show statistically significant differences between two states for each linguistic cue (see bars in Figure~\ref{fig:boxplots}, $alpha=0.05$). Median information value differs between Partial Understanding and Misunderstanding ($p = 0.032$), median gaze entropy differs between Non-Understanding and Misunderstanding ($p = 0.036$), and median syntactic complexity score differs between Understanding and Partial Understanding ($p = 0.042$).

It is worth noting that we do not assume a direct link between potentially higher cognitive load, as measured by average information value, and misunderstanding or non-understanding. However, based on our observations from Figure~\ref{fig:boxplots}, when explainees are exposed to speech which takes a higher cognitive load to process (i.e., a higher average information value and higher average gaze entropy value), they are more likely to report, in the subsequent video recall task, that they understand, or partially understand the explanation. Specifically, Figure~\ref{fig:boxplots} suggests that Understanding and Partial Understanding tendencies are slightly more prevalent when information value is higher. This may indicate that listeners have a higher probability of understanding the content when exposed to speech with a higher information value (high cognitive load). One possible explanation is that listeners are more attentive when the information value is higher which aligns with previous studies assuming that high information density may help to maintain listener's attention (see \citealt[]{tsipidi-etal-2024-surprise}, citing \citealt[]{BjareLW24} for findings on music: listener engagement can be influenced by modulating surprisal). However, this assumption still requires further experimentation. Additionally, average gaze entropy also shows a similar pattern, with higher average values for Understanding and Partial Understanding compared to the Non-Understanding and Misunderstanding. We hypothesise that higher average gaze entropy values correspond to greater variation in gaze labels, which is considered a cue to cognitive load in some empirical analyses \citep[e.g.,][]{morency2006recognizing, Glenbergetal1998}. 

In the following analysis, we go beyond individual linguistic cues and will analyse how they can be jointly used in a classification task.

\subsection{Classification Experiments}
\label{sec:classification}

We conduct classification experiments to investigate, based on the analysis of the classification result, how predictable different understanding labels. Based on the results of the statistical analysis in Section~\ref{sec:results}, we decided to only use average information value, average gaze entropy value, and syntactic complexity score, excluding average dependency length. 

\begin{figure*}
    \centering
    \includegraphics{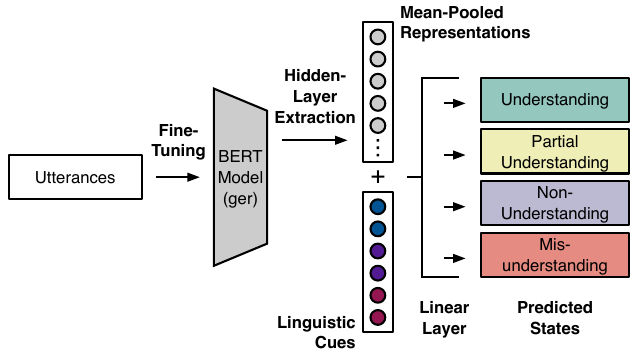}
    \caption{%
        Understanding state classification by fusing linguistic cues to a fine-tuned BERT model. We first fine-tuned a (German) BERT model with the dialogue data from the MUNDEX corpus in order to learn potential textual features related to understanding states. We then focused on the last four hidden layers, fusing them with the three significant linguistic cues identified in Section~\ref{sec:stat} (average information value, average gaze entropy, syntactic complexity score) .}
    \label{fig:multimodalbert}
\end{figure*}

We employ two commonly used classifiers, Random Forest from the scikit-learn toolkit \citep{pedregosa2011scikit} and XGBoost \citep{ChenGuestrin2016}. In addition, we also fine-tune a German BERT model \citep{devlin-etal-2019-bert, dbmdz_bert_german_2019} with the utterance data and then fuse the three selected linguistic cues into the model as a third classifier (see Figure~\ref{fig:multimodalbert}). Based on earlier empirical findings that BERT tends to encode semantic and co-reference information (which is potentially related to understanding in our study) in the higher layers \citep[see][]{tenney2018what}, we only used the last four hidden layers in the fusion step. The three linguistic cues are concatenated with the textual features represented in the hidden layer, and further passed through the linear layer to perform the classification task.\footnote{%
    The training employed 10-fold stratified cross-va\-li\-da\-tion with AdamW: 15 epochs using Cross Entropy loss; learning rate: 2$e$-5; batch size: 8; learning dropout rate: 0.2.
} To make the comparison fair, we test the performance of the three classifiers under two different settings: (1) using only textual features to predict understanding states and (2) combining textual features and the selected linguistic cues. The textual features are encoded using scikit-learn's TF-IDF for the Random Forest and XGBoost classifiers, whereas the custom BERT model encodes them as a language model.
For model evaluation, the training and testing data are split 7:3. Evaluation is performed using only the test data. In addition, 10-fold cross-validation was used as another evaluation method. The classification results are reported in Table~\ref{tab:classfication-results}. We report precision, recall, F1 score, as well as average accuracy and Macro F1 for each model accordingly.

\begin{table*}
\small
\begin{tabularx}{\linewidth}{llXXXXXXXXX}
\toprule
   & & \multicolumn{3}{c}{\textbf{Random Forest}} & 
        \multicolumn{3}{c}{\textbf{XGBoost}} & 
        \multicolumn{3}{c}{\textbf{German BERT}} \\
\cmidrule(lr){3-5} \cmidrule(lr){6-8} \cmidrule(lr){9-11}
    &\textbf{Class} & Precis. & Recall & F1 & Precis. & Recall & F1 & Precis. & Recall & F1 \\
\midrule
    \multirow{6}{*}{\rotatebox{90}{\scriptsize{}Textual only\phantom{Xx}}}
    & Understanding & 0.74 & 0.70 & 0.72 & 0.59 & \textbf{0.80} & 0.68 & \textbf{0.76} & \textbf{0.80} & \textbf{0.78} \\
    & Partial Understanding & 0.75 & \textbf{0.67} & 0.71 & 0.71 & 0.59 & 0.65 & \textbf{0.80} & \textbf{0.67} & \textbf{0.73} \\
    & Non-Understanding & 0.55 & 0.77 & 0.64 & \textbf{0.79} & 0.65 & 0.71 & 0.71 & \textbf{0.91} & \textbf{0.80} \\
    & Misunderstanding & 0.67 & 0.33 & 0.44 & 0.54 & 0.54 & 0.54 & \textbf{0.88} & \textbf{0.58} & \textbf{0.70} \\
    \cmidrule(lr){2-11}
    & 10-CV Acc. (mean ± SD) & \multicolumn{3}{c}{0.762 ± 0.050} & \multicolumn{3}{c}{0.695 ± 0.049} & \multicolumn{3}{c}{\textbf{0.768} ± 0.139} \\
    & 10-CV Macro F1 (mean ± SD) & \multicolumn{3}{c}{0.756 ± 0.058} & \multicolumn{3}{c}{0.693 ± 0.052} & \multicolumn{3}{c}{\textbf{0.758} ± 0.158} \\
\midrule
    \multirow{6}{*}{\rotatebox{90}{\scriptsize{}Textual \& Linguistic \phantom{X}}} 
    & Understanding      & 0.71 & 0.67 & 0.69 & 0.50 & 0.72 & 0.59 & \textbf{0.74} & \textbf{0.85} & \textbf{0.79} \\
    & Partial Understanding & 0.87 & \textbf{0.81} & \textbf{0.84} & 0.77 & 0.62 & 0.69 & \textbf{0.88} & 0.78 & 0.82 \\
    & Non-Understanding & 0.70 & \textbf{0.84} & 0.76 & 0.74 & 0.74 & 0.74 & \textbf{0.82} & 0.82 & \textbf{0.82} \\
    & Misunderstanding   & 0.78 & 0.64 & 0.70 & \textbf{0.83} & 0.45 & 0.59 & 0.82 & \textbf{0.75} & \textbf{0.78} \\
    \cmidrule(lr){2-11}
    & 10-CV Acc. (mean ± SD) & \multicolumn{3}{c}{0.790 ± 0.032} & \multicolumn{3}{c}{0.709 ± 0.051} & \multicolumn{3}{c}{\textbf{0.816} ± 0.038} \\
    & 10-CV Macro F1 (mean ± SD)& \multicolumn{3}{c}{0.793 ± 0.031} & \multicolumn{3}{c}{0.708 ± 0.050} & \multicolumn{3}{c}{\textbf{0.812} ± 0.042} \\
\bottomrule
\end{tabularx}
\caption{%
    Comparison of the three classifiers with textual features only (upper half) and textual features and linguistic cues combined (lower half). Best scores in bold.}
\label{tab:classfication-results}
\end{table*}

Since the classification task contains four labels, a random classifier would achieve a chance baseline accuracy of $0.25$.
First, all of the classifiers under both settings perform better than chance. 
Secondly, although the linguistic cues' effect size is small (Table~\ref{tab:kruskal-wallis-stats} in Section~\ref{sec:stat}), classification performance is better when the they are used in addition to the textual cues -- which indicates the combined effectiveness of the linguistic cues in predicting different states of understanding. 
Third, it appears that the labels Partial Understanding and Misunderstanding are harder to predict. Although for the BERT-based classifier, the F1 score for Misunderstanding is $0.78$ when we use both textual features and linguistic cues, it is still lower than predictions of the other three labels. 
Fourth, Non-Understanding appears to be the easiest label to predict. This is most apparent for the BERT based classifiers, where F1 scores are $0.8$ and $0.82$ respectively, 10 pp and 4 pp higher than for the label Misunderstanding.

\begin{figure}
    \centering
    \includegraphics[width=\linewidth]{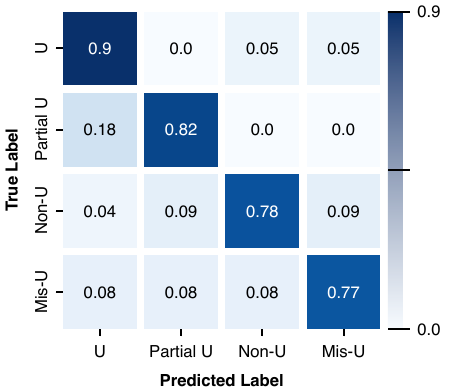}
    \caption{%
        Confusion matrix for the German BERT model for classifying understanding states (‘U’) with textual features and linguistic cues.}
    \label{fig:bert_conf}
\end{figure}

We now consider why Misunderstanding seems to be harder to predict correctly. First, we believe the weak performance to be related to the class imbalance of understanding labels. There is less training data for Misunderstanding and more training data for Non-Understanding (see Table~\ref{tab:understanding-labels}). Secondly, zooming in on the predictions made the German BERT model with textual and linguistic cues with the help of a confusion matrix (see Figure~\ref{fig:bert_conf}), it can be observed that for Misunderstanding, the model assigns uniform probability to the three other labels. In face-to-face interaction, interlocutors may not be aware at the time that their understanding is impaired, and may only later realise that they misunderstood what was said. Similar model behaviour can be seen for Partial Understanding, which tends to be predicted as Understanding.

\section{Conclusions and Future Work}
\label{sec:conclusion}

We aimed at investigating the predictability of different understanding states of explainees in explanatory interaction using different cognitive load related verbal and non-verbal linguistic cues. This builds on previous work of \citet{turk2024predictability} that analysed the predictability of different understanding states based on two classes: Understanding and Non-Understanding. For the analyses presented here, we used an updated version of the MUNDEX corpus with four different understanding states. Based on our survey of the literature, we hypothesised that the cognitive activity of understanding should be correlated with cognitive load. We chose linguistic cues potentially related to cognitive load: information value, syntactic complexity, and gaze. We used 21 explanatory dialogues of the MUNDEX corpus and quantified these three linguistic cues into four different values which may indicate cognitive load.

Our statistical analysis shows that the information value and the syntactic complexity score of explainer's utterances, as wells as explainee's gaze variation, differ significantly between listener's states of understanding. We then used these three measures in two off-the-shelf classifiers as well a fine-tuned German BERT-based classifier to conduct classification of understanding states. The results show that the BERT-based classifier generally performs much better than the two off-the-shelf classifiers, which further hints at the predictability of different states of understanding. Model performance also reveals potential challenges in predicting different understanding states, especially states labelled Misunderstanding and Partial Understanding.

As predicting states of understanding requires feature engineering of different multimodal signals, we believe there is scope to improve our current work. We consider the following to be important for interpreting different states of understanding: speech signals such as pitch and voice quality; vocal backchannels; as well as hand and head gestures. In future work, we will focus on these as additional factors. We plan to incorporate these multimodal features to see if we can improve the classification results. Another area for future research will be to try different models besides BERT to see if better classification results can be achieved.

\section{Ethical Considerations and Limitations}
\label{sec:ethics}

The publicly available MUNDEX corpus does not contain personal data of the study participants. The corpus was collected with the protection of personal data in mind, and was approved by our institutional review board.

Within the scope of this study, we consider the following three limitations which we would like to leave for future work: 
Firstly, although MUNDEX is the only publicly available multimodal corpus which provides richly annotated understanding labels, the data is still quite limited from a machine learning perspective. In the future, we hope that comparable data will be made available in order to build more reliable and robust classifiers for tracking listeners' states of understanding in human--human (and human--agent) interaction.
Secondly, the MUNDEX corpus contains data from German speakers. The results and analyses reported here therefore lack linguistic generality. It is possible that the tendency to express different states of understanding (e.g., Understanding/Non-Understanding) varies cross-culturally. In future work, we hope to use multimodal corpora of other languages in order to further generalise our findings.
Thirdly, to obtain the gaze labels, we relied on OpenFace \citep{baltrusaitis2018openface}, which could lead to the gaze labels not capturing the gaze movement with state-of-the-art accuracy (e.g., above 95\%). In future work, if more advanced gaze tracking software becomes available, we hope to create gaze labels with greater accuracy.

\section{Supplementary Material}
\label{sec:supplementary-material}

Code and data are available on Zenodo:

\noindent\href{https://doi.org/10.5281/zenodo.19003190}{https://doi.org/10.5281/zenodo.19003190}

\section{Acknowledgements}
\label{sec:acknowledgements}

Funded by the Deutsche Forschungsgemeinschaft (DFG, German Research Foundation): \href{https://gepris.dfg.de/gepris/projekt/438445824}{TRR 318/3 2026 -- 438445824}, project A02. We thank the participants and our colleagues and student assistants for their support in data collection, transcription, and annotation. We would also like to thank the anonymous LREC 2026 reviewers who provided constructive feedback to help us improve this work.

\section{Bibliographical References}
\label{sec:reference}\vspace{-0.7cm}
\bibliographystyle{lrec2026-natbib}
\bibliography{bibliography-lrec2026-understanding}

\section{Language Resource References}
\label{lr:ref}\vspace{-0.7cm}
\bibliographystylelanguageresource{lrec2026-natbib}
\bibliographylanguageresource{languageresource-lrec2026}

\end{document}